# Crossing Borders Without Crossing Boundaries: How Sociolinguistic Awareness Can Optimize User Engagement with Localized Spanish AI Models Across Hispanophone Countries

*Martin Capdevila, Esteban Villa Turek[1], Ellen Karina Chumbe Fernandez, Luis Felipe Polo Galvez, Andrea Marroquin, Rebeca Vargas Quesada, Johanna Crew, Nicole Vallejo Galarraga, Christopher Rodriguez, Diego Gutierrez, Radhi Datla*

## Abstract

Large language models are, by definition, based on language. In an effort to underscore the critical need for regional localized models, this paper examines primary differences between variants of written Spanish across Latin America and Spain, with an in-depth sociocultural and linguistic contextualization therein. We argue that these differences effectively constitute significant gaps in the quotidian use of Spanish among dialectal groups by creating sociolinguistic dissonances, to the extent that locale-sensitive AI models would play a pivotal role in bridging these divides. In doing so, this approach informs better and more efficient localization strategies that also serve to more adequately meet inclusivity goals, while securing sustainable active daily user growth in a major low-risk investment geographic area. Therefore, implementing at least the proposed five sub variants of Spanish addresses two lines of action: to foment user trust and reliance on AI language models while also demonstrating a level of cultural, historical, and sociolinguistic awareness that reflects positively on any internationalization strategy.

**Keywords**: sociolinguistics, Spanish dialectology, language localization, AI models, user engagement

## 1. Introduction

Until now, many organizations have taken a homogenizing approach to Spanish (ES) language model development throughout Latin America. The goal has been to strive towards achieving linguistic neutrality and to adhere to missions of inclusivity. Ironically, we have found this to have an opposite effect. This is not to entirely discredit previous analyses. Some linguistic analysis reports have provided detailed insights on these differences, but they have been primarily centered on differences in spoken language. However, this strategy report focuses solely on variants' differences and similarities in their written forms. Various studies show that there might be a tendency to use one translation of each language without regard to differences within the language and across cultures using that language, which is the case of Spanish, one of the most frequently spoken languages in the world (Van Zyl & Meiselman, 2015).

Furthermore, the diverse nature of linguistic traditions across Latin American borders and sociocultural boundaries is the result of a myriad of influences, including, but not limited to, Indigenous language influences, colloquial and standardized grammatical norms, and stark lexical differences. When considered alongside the historical complexities of how Spanish was originally spread by Imperial Spain's imposition of strict regulations, the present approach to Spanish model development only serves to reflect this more historical, colonizing effect on language that undermines inclusivity and ignores local linguistic differences.

[1] Corresponding author: villaturek@u.northwestern.edu

In languages such as Arabic or Chinese, their written forms are similar or even identical across regions, yet their spoken variants can differ drastically, to the point where speakers may not understand each others' dialects. By contrast, differences across pan hispanic dialects, such as the one that exists between Spain and other countries, more closely resemble English, where speakers around the world can generally understand each other seamlessly. For example English variants used in Nigeria, the US, the UK, Canada, and New Zealand are mutually intelligible. As a result, it is a universally accepted best practice that users in countries that employ enGB grammatical rules in written language, should not be served model responses with text that uses enUS grammatical rules. This is particularly true in the case of Canada, where spoken language is almost identical, but written language differs, as is the case with well-known examples like "color" vs "colour" or "center" vs "centre" (enUS vs enGB, respectively). The same is true in the opposite direction: Users located in countries that follow US grammar should not be shown answers using UK grammar or colloquial terms. Examples of this include "cell" vs "mobile", "elevator" vs "lift", or even "underwear" vs "pants" (enUS vs enGB, respectively). The latter example presents a case where slight differences in signifiers can drastically alter familiarity and local relevance.

These differences exist to a much greater degree in Spanish than they do in English. As such, there is a clear need to develop and fine-tune localized models to address the unique needs of different Spanish locales. This strategy report focuses on exploring the varieties of Spanish that share linguistic similarities or exhibit critical differences in their written forms. Given these differences, a localized approach over a homogenizing one will achieve greater daily active user (DAU) growth and engagement, improve retention and enhance overall user experience with AI models, and ultimately establish true inclusivity. Despite some vast differences across Latin American Spanish dialects, we propose a unified method for grouping them.

## 2. Definitions

Throughout this document we will refer to these terms according to the following definitions:

Pan Hispanism: Pertaining or relative to all Spanish speaking peoples in the world

Native speaker: A person who has grown up in a Spanish speaking household, and had several years of formal education in Spanish at a secondary or post-secondary level

Written grammatical characteristics: Pertaining or relative to how dialectal and formal variants of Spanish interact and differentiate among each other

Lexical differences: Pertaining or relative to the differentiated and/or unique meaning that Spanish speakers attach to words in the context of their dialect and/or location

Dialect: "a regional variety of language distinguished by features of vocabulary, grammar, and pronunciation from other regional varieties and constituting together with them a single language" (Merriam-Webster, 2025)

Signifier: Any word that is tied to a particular object or meaning

## 3. Background on Dialectal Groups

Due to complex colonialist histories shared among some widely spoken European languages, localization presents a unique challenge. For Spanish and French, the written forms of their European variants are codified and regulated by their respective central bodies founded in the early 17th century, and still presently based in Europe. Both The French Academy (L'Académie Française) and The Royal Academy of Spanish (RAE) act as singular sources for the official codification of language.

However, because regional linguistic differences in every language are rooted in speech, inter-cultural exchange, and formalized region-specific grammatical rules, even if users could understand a standardized version of European Spanish, it does not mean that using such a version across AI products would translate into familiar and seamless user experiences. The mechanism operating here is very similar to the aforementioned analogy regarding the use of enGB grammar in the enUS locale. In short, one of the key implications of such differences is that language models should be able to identify a user's locale by means of metadata (such as user's location), coupled with linguistic-based inferences according to the user's input. This "sound like me" approach is particularly true for growth-related strategies, including, but not limited to, Agentic AI applications in heavily localized contexts.

To be sure, one of the core pillars of any language model boils down to natural language understanding. Simply put: as a first step, a large language model should first understand any user input before providing an answer. In Spanish, it has been assumed that understanding users is made simpler by the linguistic foundation codified by the RAE. However, this poses an overly simplistic and altogether problematic solution, because the RAE acts as a monolithic linguistic authority. It can be viewed as having a similar role in Spanish as the Oxford Dictionary has in English worldwide. While certain basic linguistic structures and definitions are enshrined there, it simply cannot be expected to oversee the organic nature of a living language. That is, like all languages, Spanish is constantly in flux, and collective understanding is predicated on collective agreement about the respective meanings of their signifiers (words). So even if the RAE strives towards the continuation of its own linguistic authority (self-appointed and perceived), it cannot practically, or meaningfully assert active control over the evolution of Spanish across all of Latin America. Furthermore, assuming it can, should, or actually does, risks promoting a form of linguistic erasure. This is why it is essential to understand how the dialectal groups differ, what influences these differences, and how we can group them to create models that acknowledge them.

In its written form, Spanish is internationally recognized and understood, even despite differences between spoken variations. Dialectal variation comprises three major elements: phonological, lexical, and grammatical. Several linguistics-related attempts have been made to categorize and group similar and dissimilar dialects based for instance on morphology or phonology, although not one solution has been universally accepted. Figure 1 shows a world map with an attempted categorization based on such differences. Figure 2, in contrast, denotes countries in which morphology plays a distinctive role in Spanish dialects aggregation, based mainly on profound differences in terms of pronoun usage and verb conjugations, according to the Association of Spanish Language Academies. Countries marked in darker colors denote critical divergences from "standard" Spanish, such as the use of vos instead of tú, or conjugations like tenés instead of tienes, which are explored in more detail in the following sections.

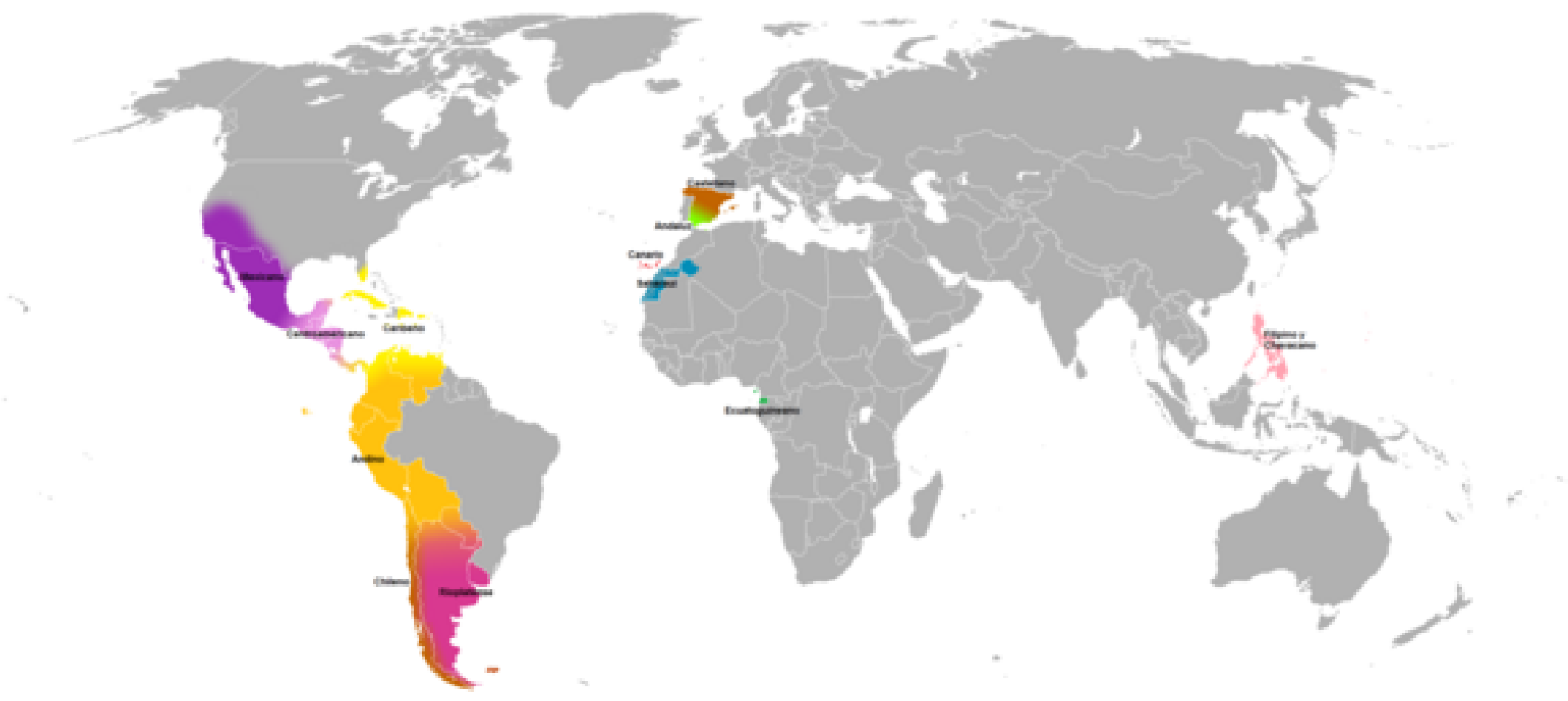

*Figure 1 - Phonological Grouping of Spanish Dialects. Adapted from "Spanish Language," Wikipedia, 2024.*

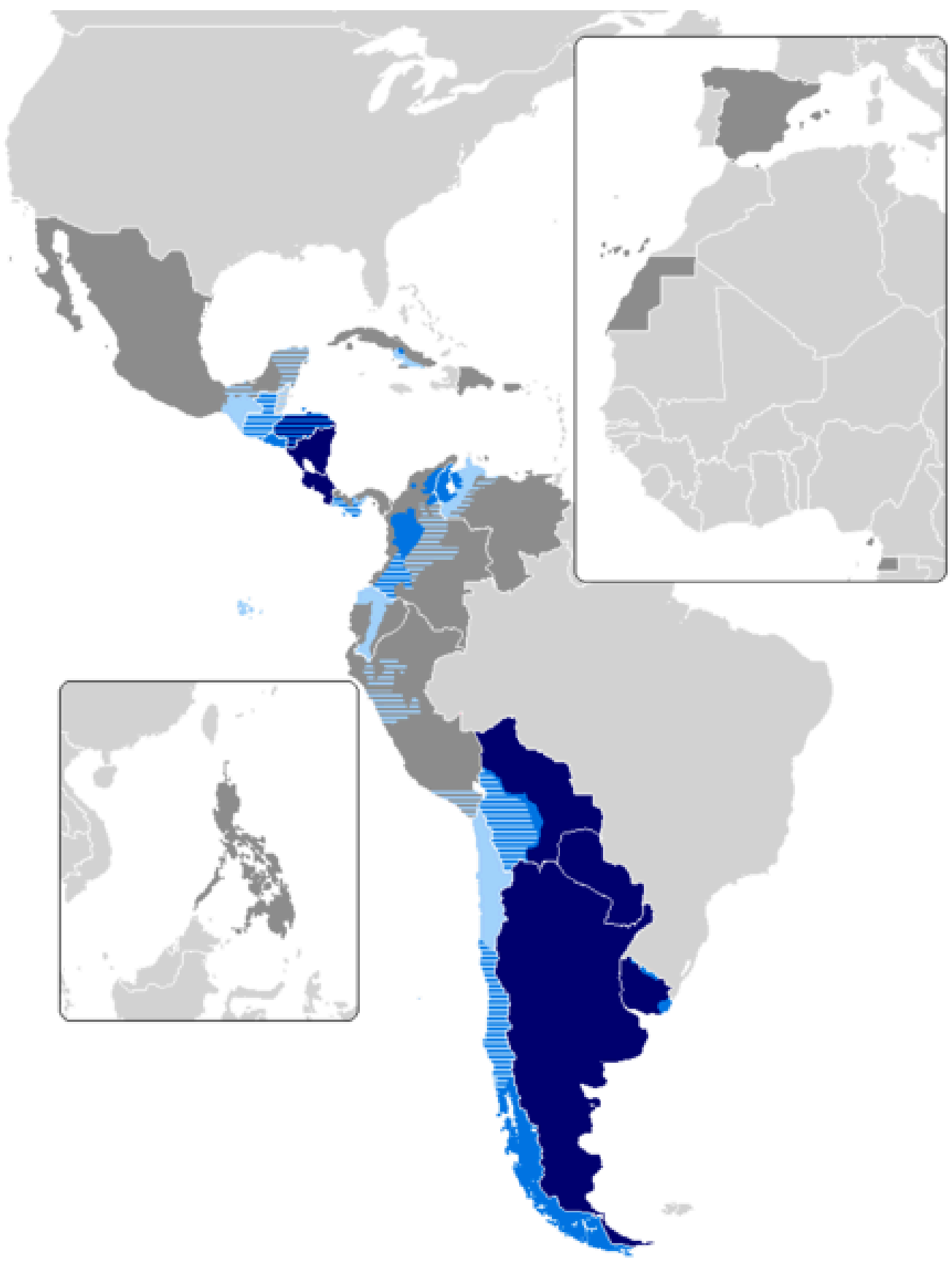

*Figure 2 - Morphological Grouping of Spanish Dialects Adapted from "Spanish Language," Wikipedia, 2024.*

However, more recent research in this area has adopted an empirical approach based on digital trace data, including publicly available social media posts. By calculating lexical usage distributions (Figures 3 and 4) and semantic-based similarities, it becomes clearer than Spanish spoken in Spain and most Latin American countries cannot be grouped and differentiated based on data alone either (see Figures 5 and 6) (Tellez et al., 2021).

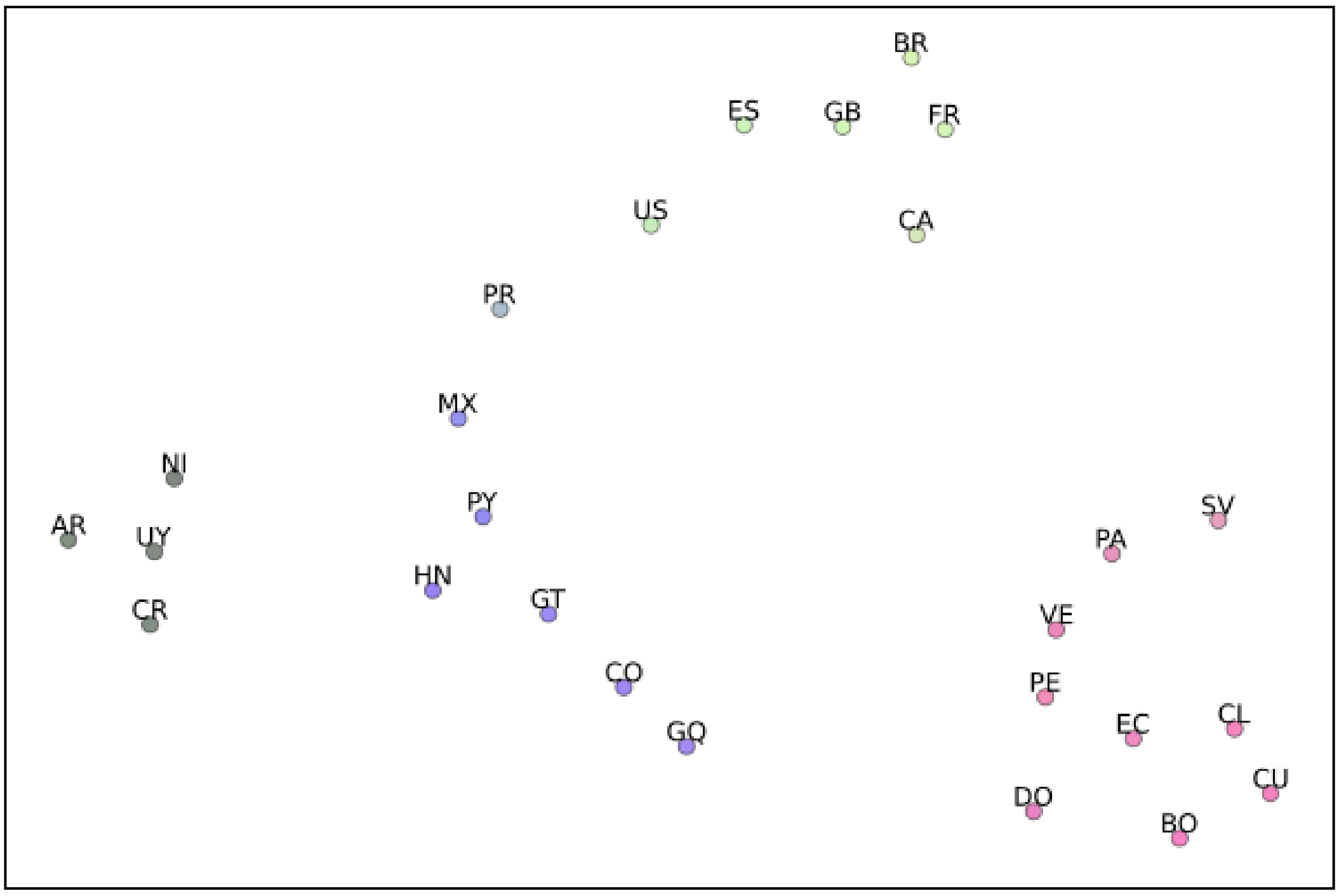

*Figure 3 - Lexical Grouping of Spanish Dialects. Adapted from Tellez et al., 2021.*

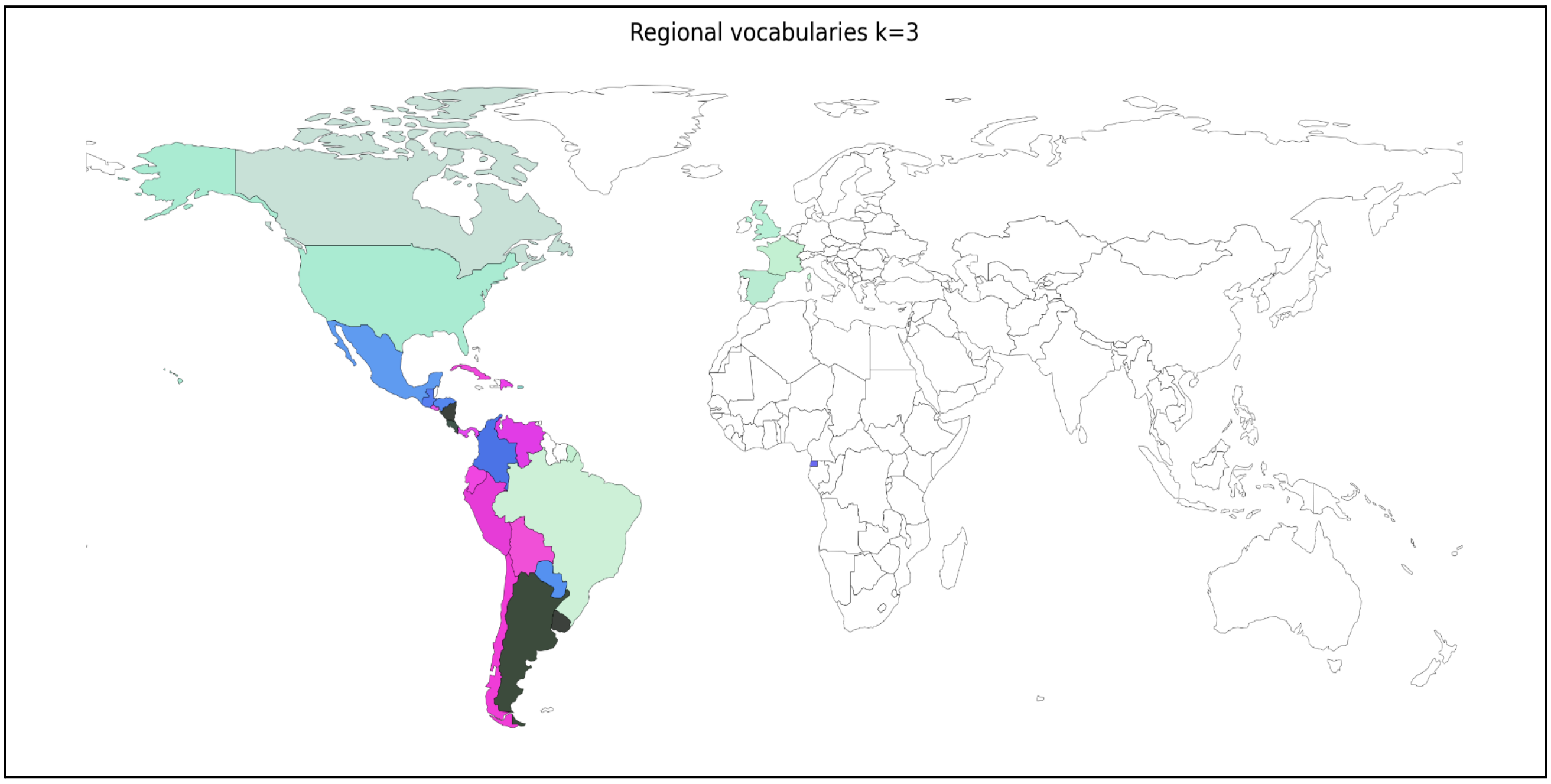

*Figure 4 - Lexical Projection of Spanish Dialects. Adapted from Tellez et al., 2021.*

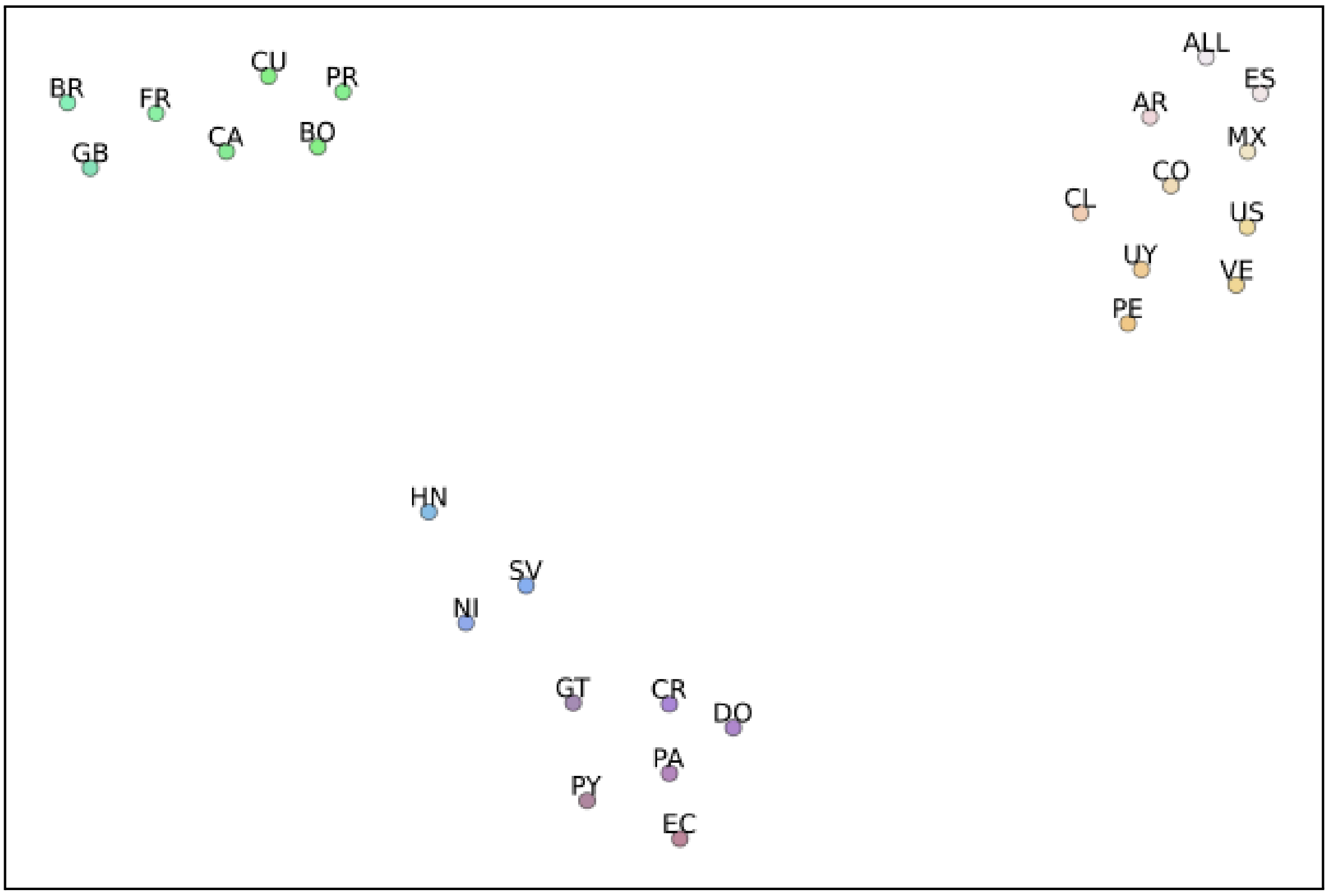

*Figure 5 - Semantic Grouping of Spanish Dialects. Adapted from Tellez et al., 2021.*

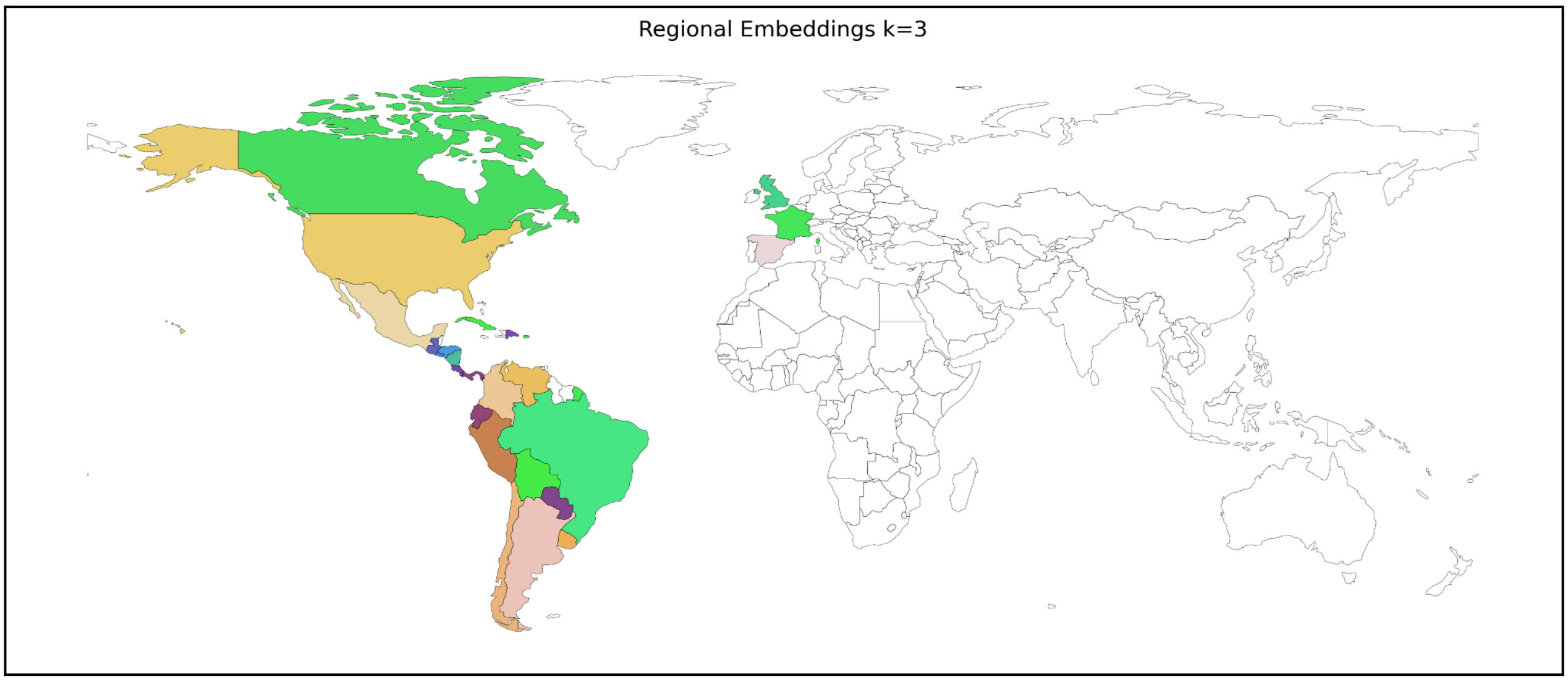

*Figure 6 - Semantic Projection of Spanish Dialects. Adapted from Tellez et al., 2021.*

# 4. Sociocultural Influences on Spanish

## 4.1 Indigenous Languages

Academic inquiry into languages has increasingly focused on their origins, development, and transculturality, and Latin America offers a unique mosaic of intercultural influence. That is because this region has seen culturally-influenced language exchange between Spanish and countless Indigenous languages for over 500 years. When Spanish colonizers arrived in the New World, the most notable Indigenous language was Nahuatl, the language of the Aztecs, whose settlements extended from present day Mexico, New Mexico, and Arizona all the way down into Central America. Other languages included Mayan, which was used in Guatemala and Yucatan, and Taíno (a dialect of Arawak) used in the Caribbean and on the coasts of Venezuela and Colombia. Quechua was the language of the Inca Empire, which encompassed the Andean countries of Peru, Ecuador, Colombia, northern Chile, and Argentina. Aymara was spoken in southern Peru, Bolivia, parts of Chile, and Argentina. Mapudungun of the Mapuche people was spoken in Chile and western Argentina, while Guaraní was spoken in Paraguay, northeastern Argentina, and parts of Brazil (Choi et al., 2024).

The below map (Figure 7) shows where major Indigenous communities are located throughout the region (Rojas, 2022). There is significant overlap between these and previous figures of Spanish variants, indicating where there has been a higher historical intercultural and linguistic exchange.

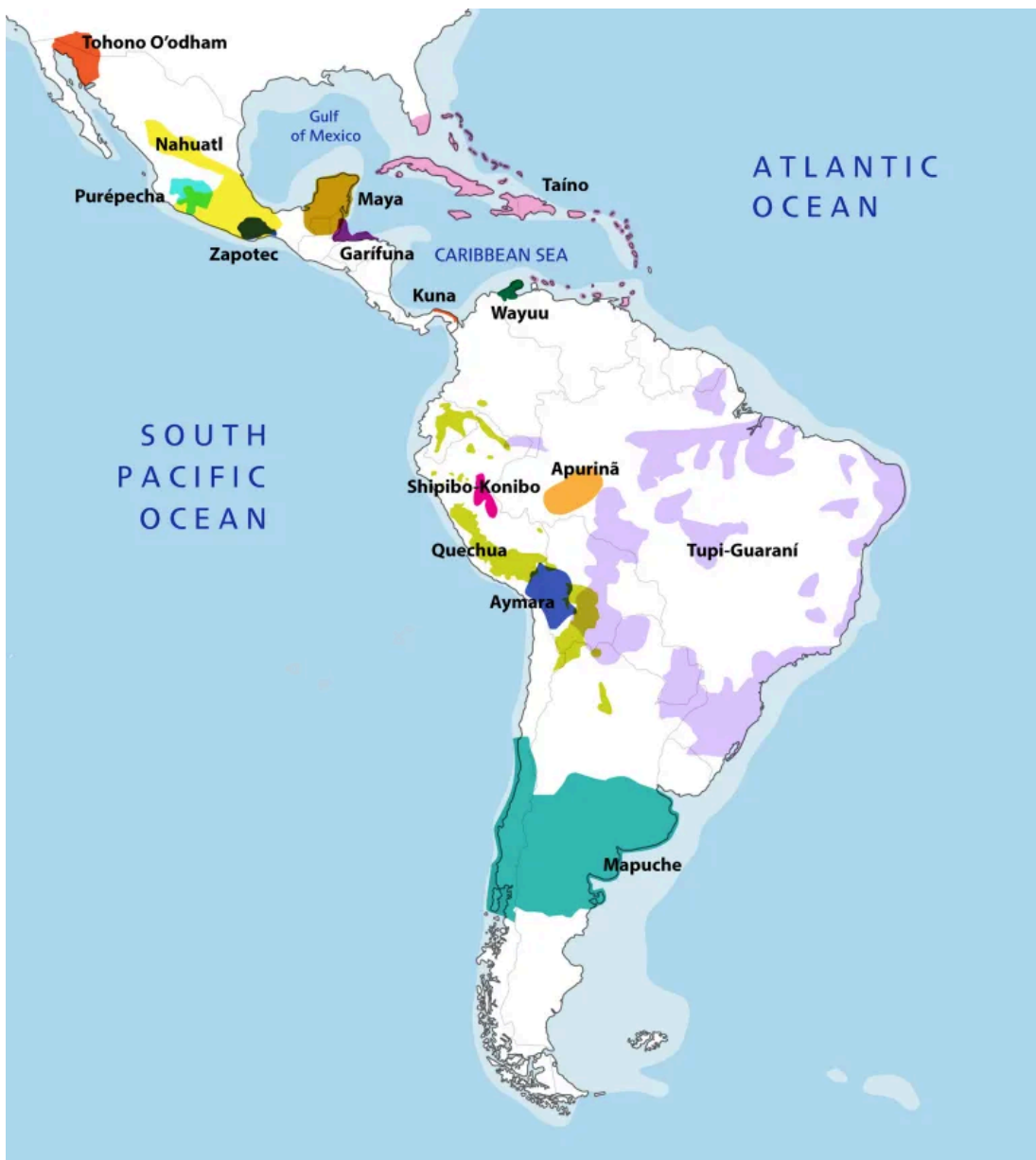

*Figure 7 - Spread of Major Indigenous Communities across Latin America. Adapted from Rojas, 2022.*

In fact, according to the Regional Observatory of Indigenous Peoples' Rights (ORDPI), the most widely spoken Indigenous language in Latin America is Quechua, with more than 10 million speakers. According to the latest census by the National Institute of Statistics and Informatics (INEI) of Peru, more than 4 million people speak an Indigenous language. Some 3,375,682 people speak Quechua, which accounts for 13.9% of the total population. Official data from Bolivia indicate that out of a total of just over 10 million inhabitants, 28% of Bolivians speak Quechua, especially in the highlands, central and southern regions of the country (Olaya Mendoza, 2024).

Today, some 550 Indigenous languages survive throughout Latin America . The most widely spoken Amerindian languages are Nahuatl in Mexico, the Mayan languages in Central America and Yucatán, Quechua in the Andes, Guaraní in Paraguay and the Río de la Plata region, and Mapudungun in Chile. These areas are characterized by daily linguistic exchanges between Spanish and Indigenous languages. Although it is true that there is no direct relationship between Quechua and the Mayan languages, a few Quechua words sound like some words in Mayan languages. It has not been determined whether they come from the same language branch, but their phonetics and semantics are somewhat similar.

Figure 8 below shows the number of people who speak the most widespread Indigenous languages across Latin America . There are between 440,000 and 770,000 Mapudungun speakers, while Nahuatl has between 1.3 and 1.7 million speakers. Additionally, between 1.6 and 2.2 million people speak Aymara, and the Mayan languages have between 4 and 5 million speakers. However, Quechuan languages boast between 7 and 10 million speakers, making it the most widely Indigenous language currently spoken today.

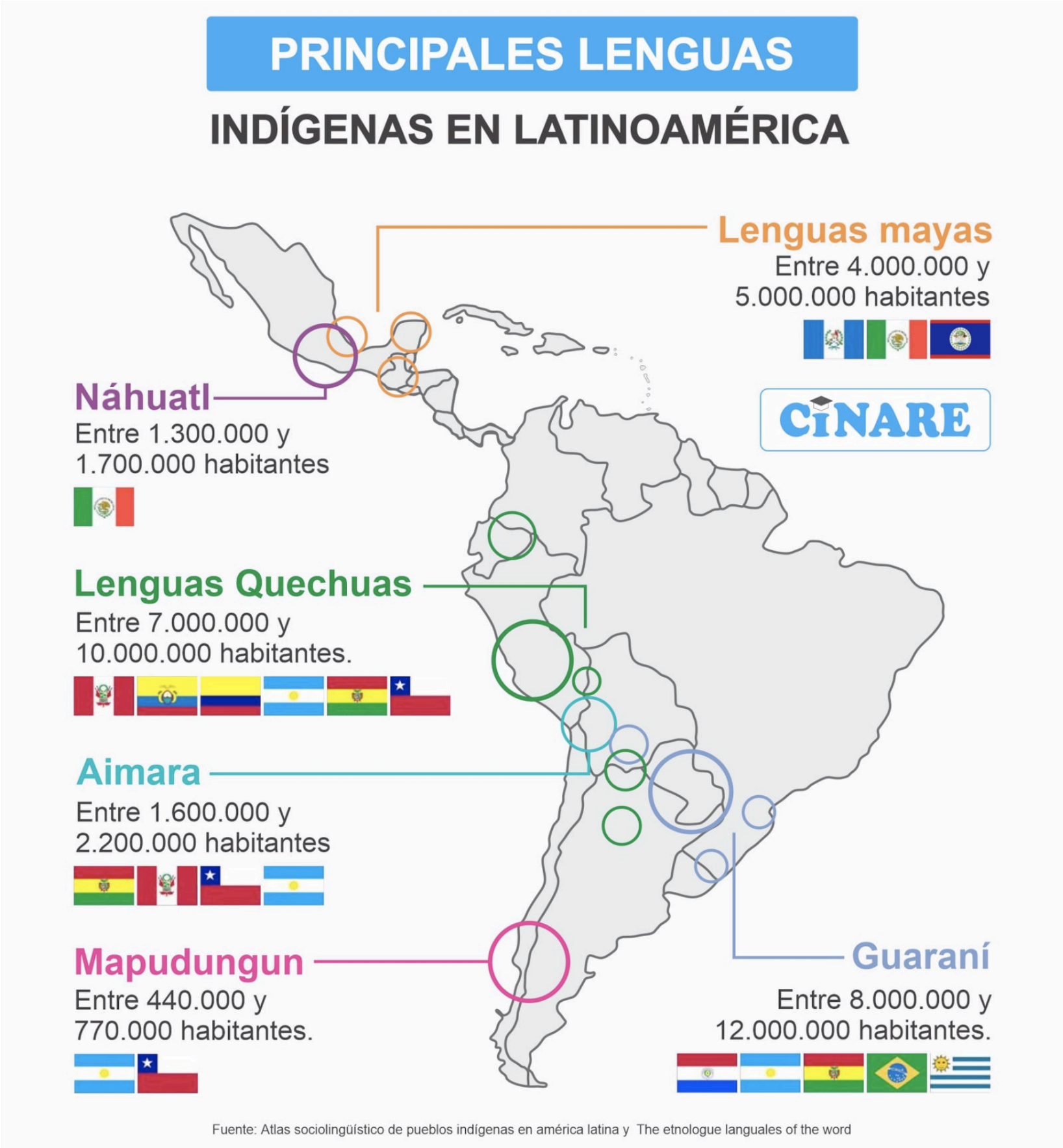

*Figure 8 - "Main Indigenous Languages in Latin America". Adapted from Choi et al., 2024.*

The Andean region demonstrates a clear example of intercultural exchange. In this area, Quechua is the most spoken Indigenous language (Caswell, 2022). According to the language database Ethnologue (2022) and Wikipedia (2024), there are about 7.2 million Quechua native speakers throughout the central Andes Mountains, including Argentina, Ecuador, Bolivia, Chile, Colombia, and Peru. Américo Mendoza-Mori, lecturer in Latinx Studies and faculty director of the Latinx Studies Working Group at Harvard University, estimates there are 8 million Quechua speakers throughout the Andes. In fact, according to a Google press release from May 22, 2022 when the company announced that a new Google Translate update would include Quechua, that number rose to 10 million speakers when Peru, Bolivia, Ecuador, and surrounding countries were included (Caswell, 2022).

In these countries Spanish coexists with Indigenous languages, creating further intercultural influence, particularly where they are designated as official languages (including Bolivia and Paraguay). The result is that colloquial Spanish itself has evolved to include Indigenous words, even in cases where users may be monolingual Spanish speakers. As such, it is difficult to deny the influence of Indigenous languages on the evolution of Spanish throughout Latin America and there is perhaps no better example of this than Quechua.

As we will explain below, many Spanish words used in this region have been heavily influenced by Quechua and have subsequently entrenched themselves in the vocabulary of primary Spanish speakers. However, the use and understanding of such words can vary drastically. One example is the various signifiers used for "corn". Given the fact that the cultural importance of corn in South American cuisine is rooted in its historical connection to Indigenous societies, Indigenous linguistic influence over its signifiers is much more prevalent as compared to Spain, where only maíz is used. What is particularly poignant here is that these words also have sub variants according to each region. Below are some of the foremost examples of Quechuan loanwords.

1. **Ñaño**: In Bolivia, Ecuador or Peru this word is used to describe having a close bond with a friend much like a sibling. However, in other countries like Colombia or Panama, it means when an individual is spoiled, which changes the meaning completely.
    a. Example: *Somos ñaños; nos criamos juntos desde que nacimos* (We're like brothers. We grew up together since we were born.)
    b. Example: *Al niño lo tienen tan ñaño que ni siquiera sabe amarrarse los zapatos*. (The child is so spoiled that he doesn't even know how to tie his shoes.)
2. **Choclo**: In most Hispanic countries, this word means corn. However, in Mexico and parts of Central America, the word changes to elote or maíz.
    a. Example: *En la sopa también puse trozos de choclo*. (In the soup I also put pieces of corn.)
3. **Chapa**: This word comes from the Quechua Chapak which means watchman. Although this is the root of the word, the meaning heavily varies with each country. For example, in Ecuador it is used as the name of a lock, but also used as a slang term to refer to "cop"; in Peru, depending on the region, the word could mean "bottle cap" or define the term "nickname". Moreover, in some countries in Central America this same word means denture.
    a. Example from Ecuador and Peru: *La puerta de la tienda tenía una chapa dorada.* (The door of the shop had a gold plate.)
    b. Example from Ecuador: *Ese chapa me detuvo.* (The cop stopped me.)
    c. Example from Peru: *Destapa la chapa y tómate la gaseosa.* (Open the cap and drink the soda.)
    d. Example from Costa Rica: *Se le cayó la chapa de la boca.* (The veneer fell out of his mouth.)

In essence, the above serves to highlight a single core principle: that the homogenization and subsequent neutralization of Spanish fails to effectively engage users by adhering to the "sound like me" approach. Indeed, Spanish is the second-most spoken language in the world with over 486 million first-language speakers (Wikipedia, 2025), and the official national language in more countries than any other language in the world. As a result, the key starting point for a localized Spanish model would be to focus, at least initially, on comprehension of user inputs, while simultaneously striving to optimize the core linguistic foundation of Spanish.

However, it must be underscored that the above only serves as an example to differentiate between the dialects, it does not identify the groups needed for localization. In fact, perhaps the mere existence of these Spanish groupings points to the need for more diverse localized AI models that can successfully meet the needs of locales, while also avoiding blanketing entire regions with dialectal differences that may or may not stretch across borders. In other words, in order to cross geographical borders throughout Latin America, and at the same time avoid traversing dialectal boundaries, a different approach is needed, one that takes into consideration the influence of cross cultural exchange with Indigenous languages and the linguistic differences between Latin America countries that may significantly mark any given locale's unique variation of Spanish. These differences and influences are outlined below.

### 4.2 Pop Culture Influence

The level and diversity of Spanish used in urban Latin music can serve as a powerful tool in the development of AI models, particularly for localization. Given the popularity of the genre and widespread consumption of Latin music among Spanish speakers, particularly those of lower and middle class communities, it would be extremely useful for localized Spanish models to be familiar with the language used (including idioms, expressions, words, etc.) by artists. This is because such linguistic variations more accurately reflect the quotidian use of Spanish by the general public, and thus more accurately represent larger communities. It will also inform the framework and/or benchmark for the Latin American - Miami dialect outlined below, given that the language used in the genre is predominantly produced in the city. The following section focuses on the genre of reggaeton and its influence on the Latin American - Miami Spanish dialect.

Reggaeton has had a profound influence on Latin American pop culture and identity. Many of the genre's artists use their music to address issues that resonate with marginalized and oppressed groups. Additionally, reggaeton has contributed to the globalization of Latin American culture given its popularity beyond its borders, reaching audiences across the world. This led to increased cultural exchange within diverse speaking communities (WeChronicle, n.d.). This emphasizes the influence of reggaeton in Spanish vocabulary. Since it so heavily influences the Latin American - Miami Spanish dialect, it is clear that this variant is perhaps the most culturally interconnected and diverse variation of Spanish. In fact, Reggaeton's influence on language and vocabulary extends beyond its musical and lyrical dimensions. The genre has become a platform for linguistic diversity and cultural representation, highlighting the richness of linguistic expressions within urban and hip-hop cultures. Through its use of regional dialects, bilingualism, and code-switching, reggaeton has become a vehicle for preserving and celebrating language diversity. Furthermore, reggaeton artists often integrate elements of their cultural heritage into their lyrics, infusing their work with linguistic and cultural nuances that resonate with audiences worldwide. By doing so, these artists contribute to the continuous evolution and enrichment of urban vocabulary, fostering a sense of inclusivity and interconnectedness across diverse linguistic and cultural landscapes (Music369.com, n.d.). Thus, Latin American music should be considered as a rich source of influence on culture and language alike, and Miami its geographical confluence point.

## 5. Linguistic Differences in Spanish

### 5.1 Grammatical Variations

Grammatical usage across Spanish variants can vary significantly depending on locales. For example, there are stark differences in the use of second-person pronouns in the singular and plural forms. In Latin America, the second-person pronouns of the singular are used as both *tú* or *vos*, and these clear cut differences between pronouns necessarily dictate a differentiated verb conjugation as well. In particular, verbs that are conjugated with the *tú* pronoun would be conjugated as, for example, *tú haces* or *tú comes* ("you do" and "you eat", respectively). Similarly, verbs that are conjugated with *vos* take a different shape, as in *vos hacés* and *vos comés*. In the case of the second-person pronoun of the plural, the differences arise on the other side of the Atlantic Ocean. In Spain, it is more common to use *vosotros* in familiar and informal contexts, and *ustedes* in formal contexts. Here, again, the respective conjugations take different forms, like *vosotros hacéis* ("you all do") or *vosotros coméis* ("you all eat") in informal contexts, and *ustedes hacen* or *ustedes comen* for formal applications. However, in Latin America it is *ustedes* the only pronoun used for the second person of the plural, regardless of formal or informal contexts. Another quintessential example of the differences in grammar between Spain and Latin America pertains to the conjugation of verbs in the past tense. Spain uses exclusively the past perfect (*he hecho, or he comido*) for events that have happened within the current day, and the past simple for events before that (*hice or comí*). On the other hand, in Latin America only the past simple (*hice or comí*) is used for any and all past events, whether they immediately ended, occurred earlier in the day, or in the previous day. This very roughly correlates with the differences of past tense usage between enGB and enUS.

In addition to these conjugations, the context where AI models shall be used will impact localized responses differently. For example, in Agentic applications geared towards user interactions, model responses should not only be linguistically localized, but also adjusted according to the pragmatic use of the dialect, i.e., depending on the user and the context. For instance, this could be based on the demographic that a particular business might be targeting, such as younger, urban customers for whom an informal and familiar Spanish may be preferable. This means that the use of slang is not only acceptable, but even expected. On the other hand, if a specific context requires the use of more formal Spanish (such as for information searches for professional purposes), the differences between the second-person pronouns in the singular and plural forms would always default to *usted/ustedes* in both Latin America and Spain and, conversely, slang is to be avoided in these situations.

Another example of grammatical difference can be observed through clitic doubling. This term explains the use of pronouns "in verb phrases together with the full noun phrases that they refer to", taking into account the gender or number of the object referred (Wikipedia, 2024). This phenomenon is found in at least 11 languages, including Spanish. However, Spanish-recessive bilinguals (who speak Quechua or Aymara natively) in the Andean region and part of the Rio de la Plata region, use *lo* to double objects, indistinctly from gender or number (Lipski, 2012). This is outlined below:

- *Pasámelo el plato* (Pass me the plate) This is an example of doubling objects respecting gender and number. Without the adaptation explained before, it would be Pásame el plato (without lo)
- *¿Me lo va a traer la billetera?* (Will he/she bring me the wallet?) In this case, *lo* (masculine) refers to la billetera (feminine). Without the adaptation, it would be *¿Me va a traer la billetera?*

The above is another set of crucial linguistic differences in the Spanish language, the use of which has significant implications around in-context meaning and appropriateness. Ignoring them, as well as ignoring the lexical variations outlined above, would create sociolinguistic dissonance and alienation in users' experiences, subsequently impeding their long-term adoption of AI products.

## 5.2 Regional Lexical Differences

As a direct result of the above differences rooted in dialectal and cultural influence, Spanish demonstrates clear regional lexical differences. In the United States, for example, culture plays the most important role as an engine for lexical evolution. This is the result of migration at key points of confluence, such as Miami, where those cultures mix and engage in intercultural exchange and influence, resulting in lexical variations, which casts a spotlight on the fact that regional, everyday usage of Spanish can differ significantly across regions. The examples in Table 1 are the words used to signify "popcorn" and "straw", including but not limited to:

Table 1: Different ways to say "popcorn" and "straw"

| **Word in English** | **Word in Spanish** | **Countries** |
|---|---|---|
| Popcorn | *Cancha/Canchita* | Peru |
| | *Canguil* | Ecuador |
| | *Cotufa* | Spain, Venezuela |
| | *Crispetas* | Colombia, Mexico |
| | *Cabritas* | Chile |
| | *Maíz pira / Maíz tote* | Colombia |
| | *Millo* | Panama |
| | *Palomitas* | Chile, Costa Rica, El Salvador, Spain, Mexico, Dominican Republic |
| | *Palomitas de maíz* | Costa Rica, El Salvador, Honduras, Mexico, Nicaragua, Panama, Peru, Dominican Republic |
| | *Pipoca* | Argentina, Bolivia |
| | *Pochoclo* | Argentina, Chile |
| | *Poporopo* | Guatemala |
| | *Pop* | Uruguay |
| | *Cocaleca* | Costa Rica |
| | *Pororó* | Argentina, Paraguay, Uruguay |
| | *Rositas de maíz* | Cuba |
| Straw | *Absorbente* | Cuba |
| | *Bombilla* | Chile, Bolivia, Paraguay, Argentina, y Uruguay |
| | *Calimete* | Dominican Republic |
| | *Cañita* | Peru |
| | *Carrizo* | Panama |
| | *Pajilla* | Costa Rica, Chile, El Salvador, Honduras, Guatemala |
| | *Pajita* | Spain, Uruguay, Paraguay, Bolivia, Argentina |
| | *Pitillo* | Colombia, Venezuela |
| | *Popote* | Mexico |
| | *Sorbete* | Argentina, Peru, Dominican Republic, Ecuador |
| | *Sorbeto* | Puerto Rico |
| | *Sorbito* | Uruguay, Paraguay, Bolivia, Argentina |

## 5.3 Slang and Homonyms

While slang is ubiquitous across all languages and their variants, Spanish Poses a unique challenge precisely because of its near universal understanding among speakers, creating a plethora of homonyms. This is because there are countless words that are used on a daily basis and that can be understood by all Spanish speakers, but their use in the context of a specific variant locale can dictate what the meaning of the word actually is and whether their use has a particular application in the context of slang. In Table 2, for example the meaning of the word *amigo* (friend) can be implied using varied slang depending on the particular local dialect:

### Table 2: Slang terms for *amigo* (friend)

| **Word used in Spanish** | **Country** |
|---|---|
| *Bróder, Brother, Bro* | Peru, Ecuador |
| *Pata, Causa* | Peru |
| *Chero/a* | El Salvador |
| *Tronco, Colega* | Spain |
| *Compañero/a, compi* | Cuba, Nicaragua |
| *Carnal/a, Guey* | Mexico |
| *Cuate/a* | Mexico, Guatemala |
| *Parce, Parcero, Marica* | Colombia |
| *Boludo/a* | Argentina and parts of Uruguay |
| *Pana* | Ecuador and Venezuela |
| *Chamo/a* | Venezuela |
| *Llave* | Ecuador, Colombia, Mexico |
| *Ñaño/a* | Ecuador, Cuba |
| *Pibe* | Uruguay |
| *Bato* | Mexico and -lately- Central America |
| *Mae* | Costa Rica |
| *Bro, Brodel, Broski* | Puerto Rico |
| *Tigre* | Dominican Republic |
| *Compa* | Costa Rica |

Overall, Spanish slang demonstrates why multiple Spanish variants are necessary. Even the most basic words can have altered meanings depending on where they are used. The universality of Spanish across the world is perhaps behind this phenomenon: while nativeSpanishspeakers are able to understand slang words, it is the particular region which informs the actual meaning in context, as well as the correct and usual use of those words.

## 5.4 Morphology

Diminutive suffixes are frequently used throughout the Spanish speaking world. However, their use and styles change based on regions, variants, and even culture. These should be a significant part of localized models and will add much value to the user experience by means of a closely familiar approach to local Spanish variants. That is, a model would be adhering closer to the localized use of Spanish in order to align more closely with the cultural relevance of its use. Given that diminutives vary, they are therefore used to express a broad range of meanings with respect to the words they modify, altering meaning in relation to size, endearment, or scorn (Lipski, 2012). It is important to distinguish when they can be used, expected to be used, and the message they convey. Table 3 includes some examples of how they are applied in sentences and their meanings:

Table 3: Ways in which diminutives are used among Spanish dialects

| **Diminutive** | **Example** |
|---|---|
| *-ito/-ita* | - *¡Mira esa carterita!* (Look at that small bag! or Look at that cute bag!)<br><br>Explanation: Diminutive *-ita* applied to the noun *cartera* (bag) makes it *carterita*. This could express that the size of the bag is small or that the person likes the bag and is referring to it with endearment. |
| | - *¡Hola compadrito! ¿Qué cuentas?* (Hi my friend! How are you doing?)<br><br>Explanation: *-ito* applied to the noun *compadre* makes it *compadrito*. Because *compadr*e refers to a friend, the diminutive is expressing endearment.<br><br>One can refer to another person with this diminutive also, applying it to the other person's name, e.g.: *Carlita* (instead of the name *Carla*). This indicates a level of |
| | friendship or closeness between the interlocutors or that someone is trying to establish a closer relationship with a person. |
| | - *¿Quieres un cafecito?* (Do you want a coffee?)<br><br>Explanation: *-ito* applied to *café* makes it *cafecito*. This doesn't mean the person is offering a small-size coffee, but that they enjoy coffee and so they refer to it with endearment. Or even if the person doesn't like coffee that much, they want to be welcoming with their interlocutor. |
| *-illo/-illa* | - *¡Qué va a saber esa abogadilla!* (What does that "lawyer" know?"[2])<br><br>Explanation: *-illa* applied to *abogada* expresses disdain, mockery. |
| | - *¿Vas a salir otra vez con tu amiguillo?* (Are you going out again with that friend?)<br><br>Explanation: *-illo* applied to *amigo* and with a tone that expresses disdain means that the friend is a questionable one. |

On the other hand the diminutive suffixes *-ico/-ica* are predominantly used in Cuba and Costa Rica whereas *-ito/-ita* are more frequently used in Colombia, and occasionally elsewhere in the region. Diminutives can even be used to indicate completely different things. This can be seen with the uses of *-ito/-ita* and *-illo/- illa*. For example: *camita* (small bed) vs *camilla* (stretcher), or *mesita* (small table) vs *mesilla* (bedside table), and even *casita* (small house) vs *casilla* (simple and inexpensive house). Additionally, Spain applies different diminutive suffixes in different regions, as listed below (Lipski, 2012):

*-iño/-iña* is usual in Galicia
*-ín* in Asturias
*-ete* in Catalunya, Valencia and parts of Aragon
*-ino/-ina* is still used in Extremadura
*-ico/-ica* is still used in Judeo-Spanish, in Aragon, Navarra, and Murcia
*-iquio/-iquia* is occasionally used in Murcia and parts of Granada
*-icho/-icha* is occasionally used in Aragon, La Mancha and surrounding areas

[2] Here "Lawyer" in quotes denotes an emphasis on the disdain that this particular use of diminutive has across some ES dialects

# 6. Localized Model Variant Groups Based on Written Language

Given all of the aforementioned examples, data, and contexts, lexical and grammatical variance across Latin America and European Spanish form two pillars of marked linguistic difference. These are based on established characteristics (Lipski, 2012). Together they demonstrate a need for additional model variants, and we propose five core groupings, listed in descending order below. Each group of multiple variants includes all of the previous variants as well.

## 6.1 Using 1 Variant - Latin American - Miami Spanish

This variant uses *tú* for the second-person singular. Latin American - Miami Spanish is currently the most widely used dialect for translations and dubbing. This is especially true, given that it represents a melting pot of Spanish speakers in the region, historically formed by the influx of its three main Spanish variants, namely dialects from Colombian, Cuban and European-Peninsula national origins (Carter & Callesano, 2018). The presence of several different dialects in this region has led to the emergence of a more broadly understood Spanish variant. Moreover, this Spanish variant has been crucially adopted across all social strata in the region, making it the only case where this is true in the United States. Miami-Dade County is not only the region with the highest percentage of Hispanic inhabitants, but it is also the county with the highest percentage of Caribbean-Hispanic and South American inhabitants anywhere in the country (U.S. Census Bureau, 2023), solidifying Miami as the most diverse and nationally-representative Spanish speaking city in the United States (Carter & Lynch, 2015):

- 34.4% of city residents were of Cuban origin
- 15.8% had a Central American background (7.2% Nicaraguan, 5.8% Honduran, 1.2% Salvadoran, and 1.0% Guatemalan)
- 8.7% were of South American descent (3.2% Colombian, 1.4% Venezuelan, 1.2% Peruvian, 1.2% Argentine, 1.0% Chilean and 0.7% Ecuadorian)
- 4.0% had other Hispanic or Latino origins (0.5% Spaniard)
- 3.2% descended from Puerto Ricans
- 2.4% were Dominican
- 1.5% had Mexican ancestry (Wikipedia, 2024)

Furthermore, its central geographical location means that this variant is not only the most representative, but also the most prevalent throughout the eastern United States, including major hispanic enclaves like the New York, Washington and Chicago Metropolitan Areas. Finally, it is worth mentioning that Miami has established itself as a major popular culture powerhouse for all of Latin America in recent years, including music, cuisine and art, among many others, having been called the "Capital of Latin America" for decades (Booth, 1993). Moreover, the greater Florida region is a key global hub for tourism in the United States, which furthermore facilitates intercultural and linguistic exchange, thereby driving the evolution and widespread adoption of its unique Spanish dialect.

Figure 9 highlights the regions where variant 1 would be most commonly used and shared (Lipski, 2018). It should be noted that these can include entire countries, but as the figures demonstrate, this is not always the case. This underscores the importance of differentiating based primarily on dialects, rather than by geographic boundaries such as regions or countries.

**This group includes**: Cuba, Venezuela, Puerto Rico, Dominican Republic, Panama, Colombia, Costa Rica, Nicaragua, Guatemala, El Salvador, Honduras, Belize and coasts of Mexico. The largest United States regions by population, including the Miami–Fort Lauderdale–West Palm Beach Metropolitan Statistical Area, New York City, or the Tri-State, Houston–The Woodlands–Sugar Land, and the Chicago and Washington metropolitan areas, among others, are components of this group as well.

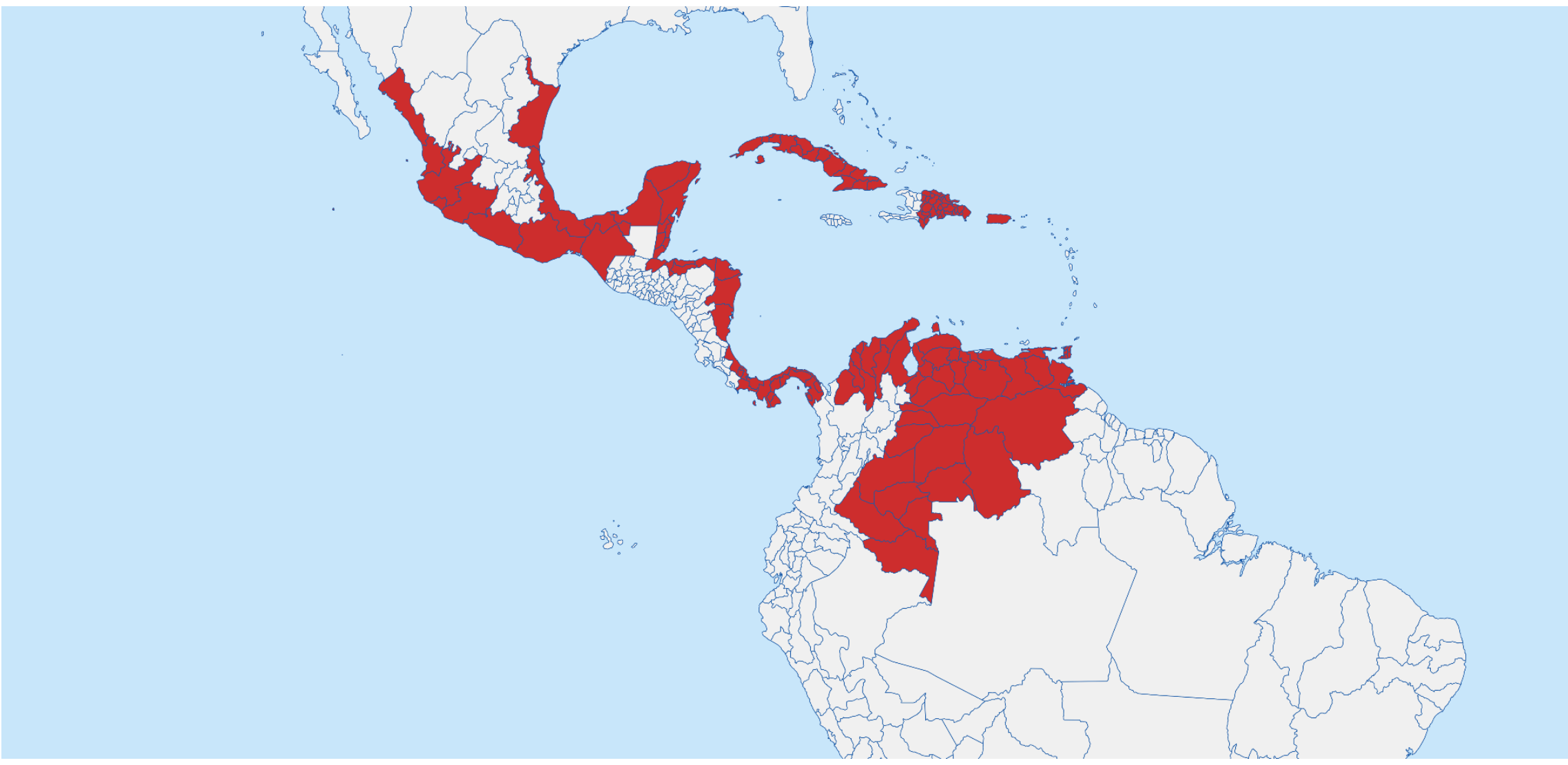

*Figure 9 - Map of Key Areas of Use for Variant 1 Across Latin America. Created by the authors.*

With respect to the aforementioned focus on inclusivity, this variant is evidence that localized models can help avoid an alienating approach to the Spanish language. The following illustrates how centralized, homogenizing strategies are problematic. Bourdieu (1984, 1991) theorized that standard language varieties, most often based on the dialect forms associated with social elites, get constructed in opposition to nonstandard varieties, and over time the linguistic preferences and patterns of usage of the elites are normalized and constructed as "natural" or "legitimate," while the forms of speech associated with non-elites are constructed as naturally and obviously inferior. In Bourdieu's view, language in late modernity is thus a form of symbolic capital that reinforces an ostensibly "natural" or "underlying" social structure in which dominant groups can effortlessly reassert their dominance, while marginal groups are systematically re-marginalized (Carter & Callesano, 2018, p. 69).

## 6.2 Using 2 Variants - European Spanish

This variant is used only in Spain, but it also forms the historical foundation for all "official" linguistic rules in Spanish, promulgated by the Royal Academy of Spanish (RAE). As mentioned above, one characteristic difference is the conjugation of verbs in past tense, with the use of the past perfect, compared to the past simple in Latin America, or the differentiated forms of second-person pronouns of the plural (*vosotros* or *ustedes*), depending on the formality of the context.

**This group includes**: Spain, Equatorial Guinea, and other European nations that have Spanish immigrants.

This variant can be understood by all regions (in its majority except for certain expressions) but is not necessarily liked.

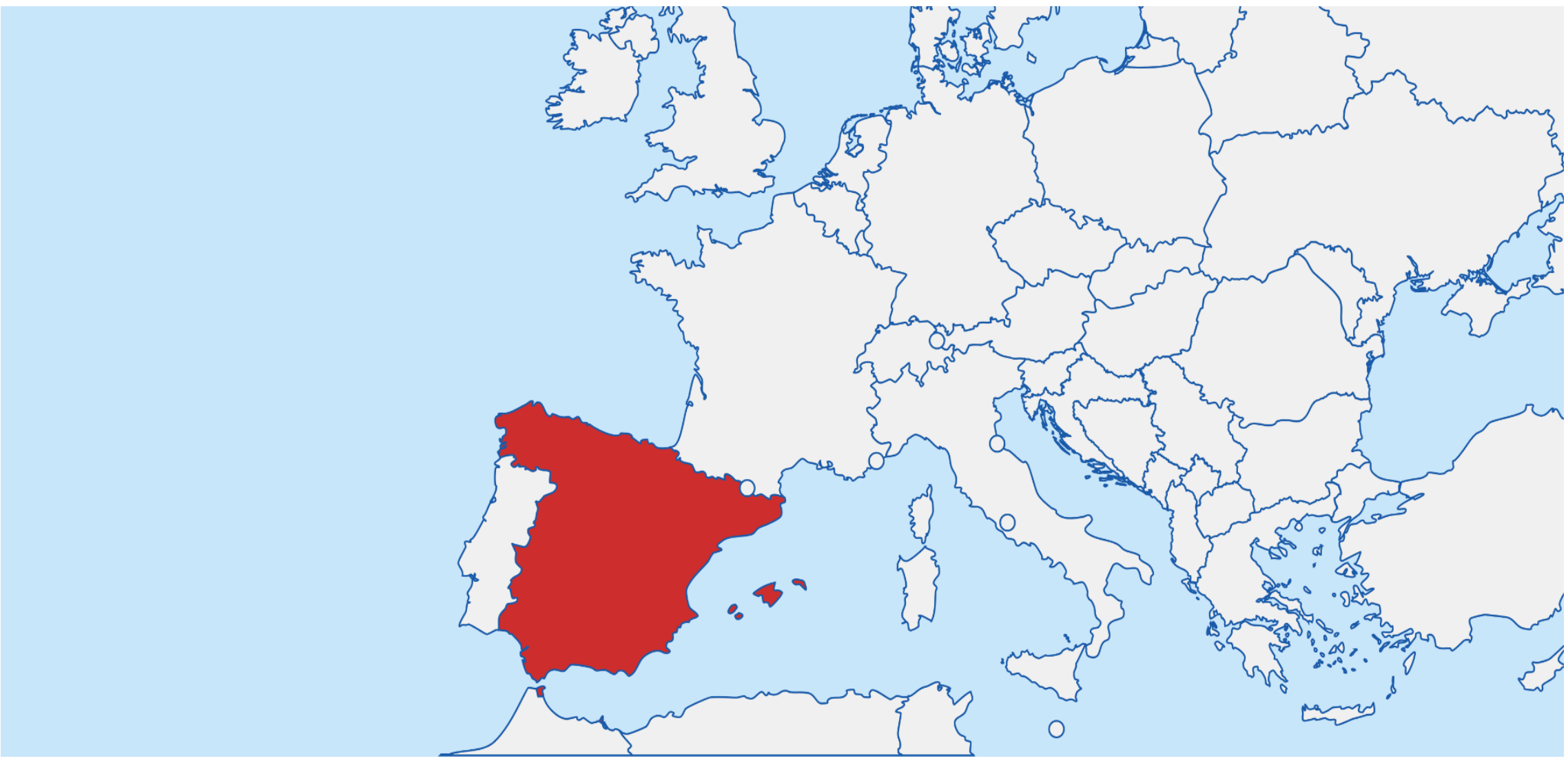

*Figure 10 - Variant 2 is used almost exclusively in Spain. Created by the authors.*

## 6.3 Using 3 Variants - Mexican Spanish

This variant is not only spoken in Mexico. It is also the most prevalent form of Spanish spoken across significant Hispanic communities located in the southwest of the United States and along the Mexican border. We must keep in mind that there are some regional variations within the country as well.

**This group includes**: Inland Mexico and Baja California, California, Texas, Arizona, New Mexico and along the border.

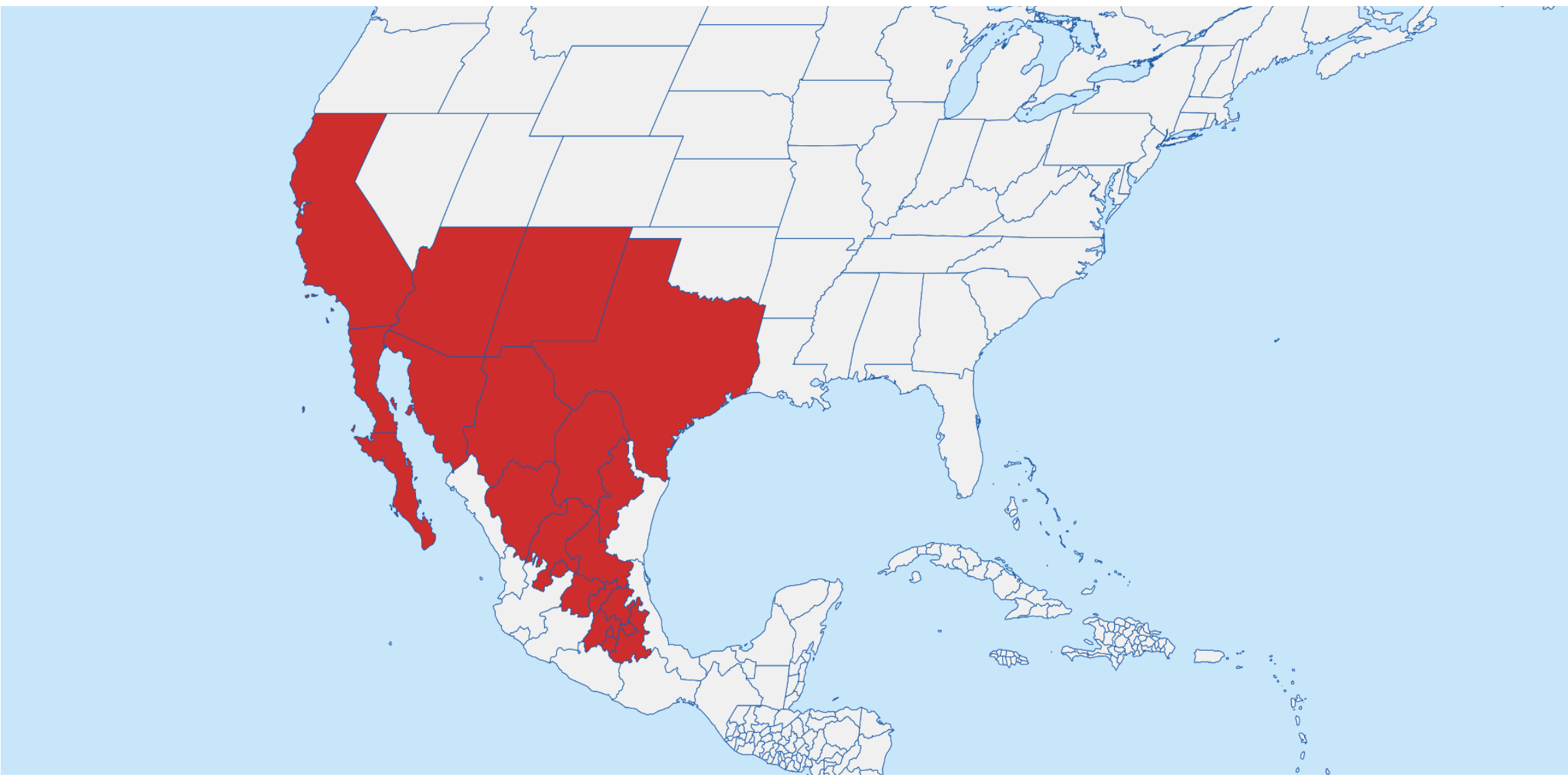

*Figure 11 - Map of Key Areas of Use for Variant 3. Created by the authors.*

## 6.4 Using 4 Variants - Rioplatense Spanish

These are countries that primarily use *vos* for second-person singular. It is predominant in South Cone countries (except Chile), and Central America.

**This group includes**: Argentina, Uruguay, Paraguay, Guatemala, El Salvador, Honduras, Nicaragua, Costa Rica.

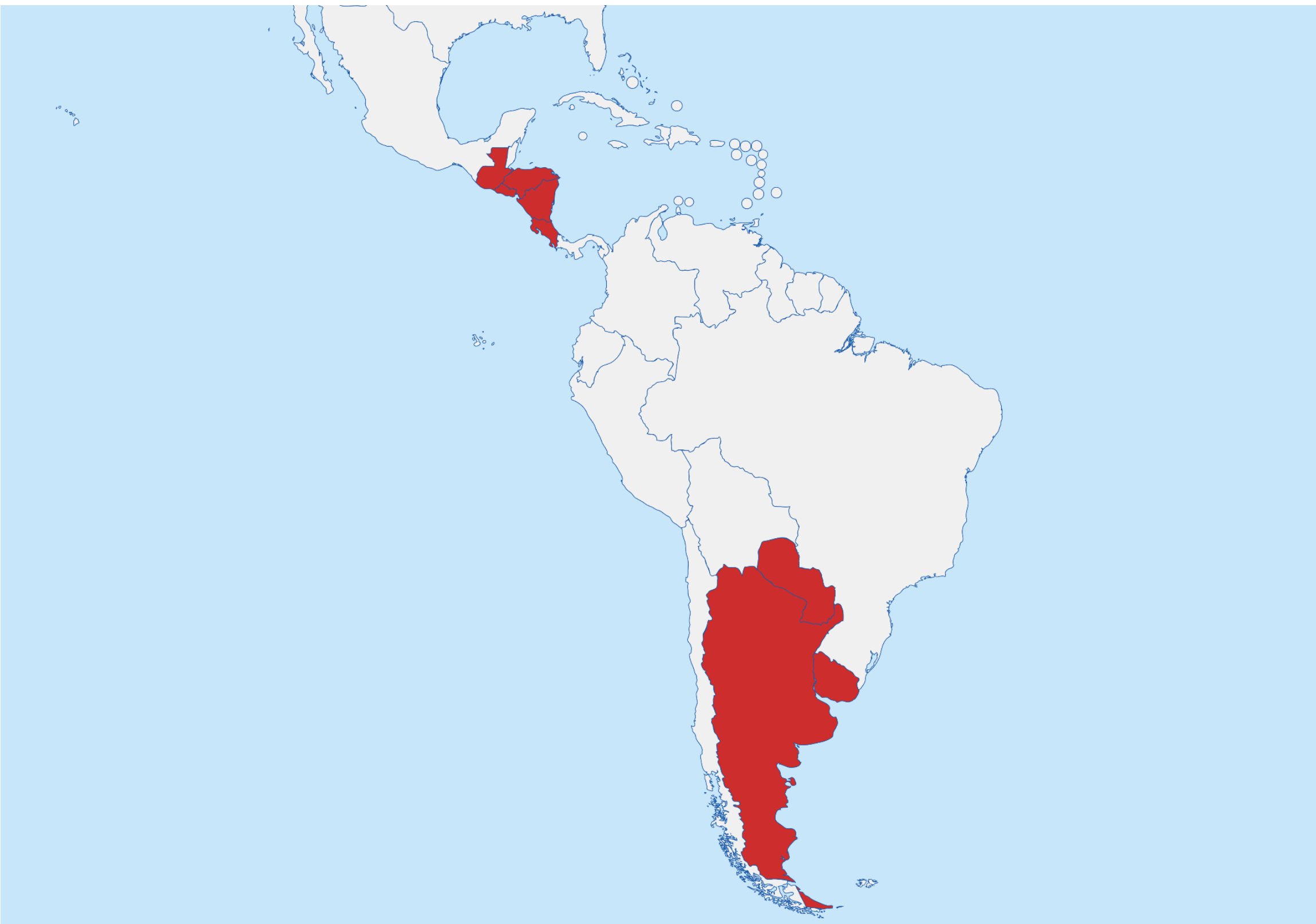

*Figure 12 - Map of Key Areas of Use for Variant 4. Created by the authors.*

## 6.5 Using 5 Variants - Andean Spanish

These are countries that also primarily use *tú* for second-person singular. This is also the variant that has the highest level of linguistic exchange with Indigenous languages.

**This group includes**: Panama, Southern Colombia, Venezuela, Ecuador, Peru, Bolivia, and Chile.

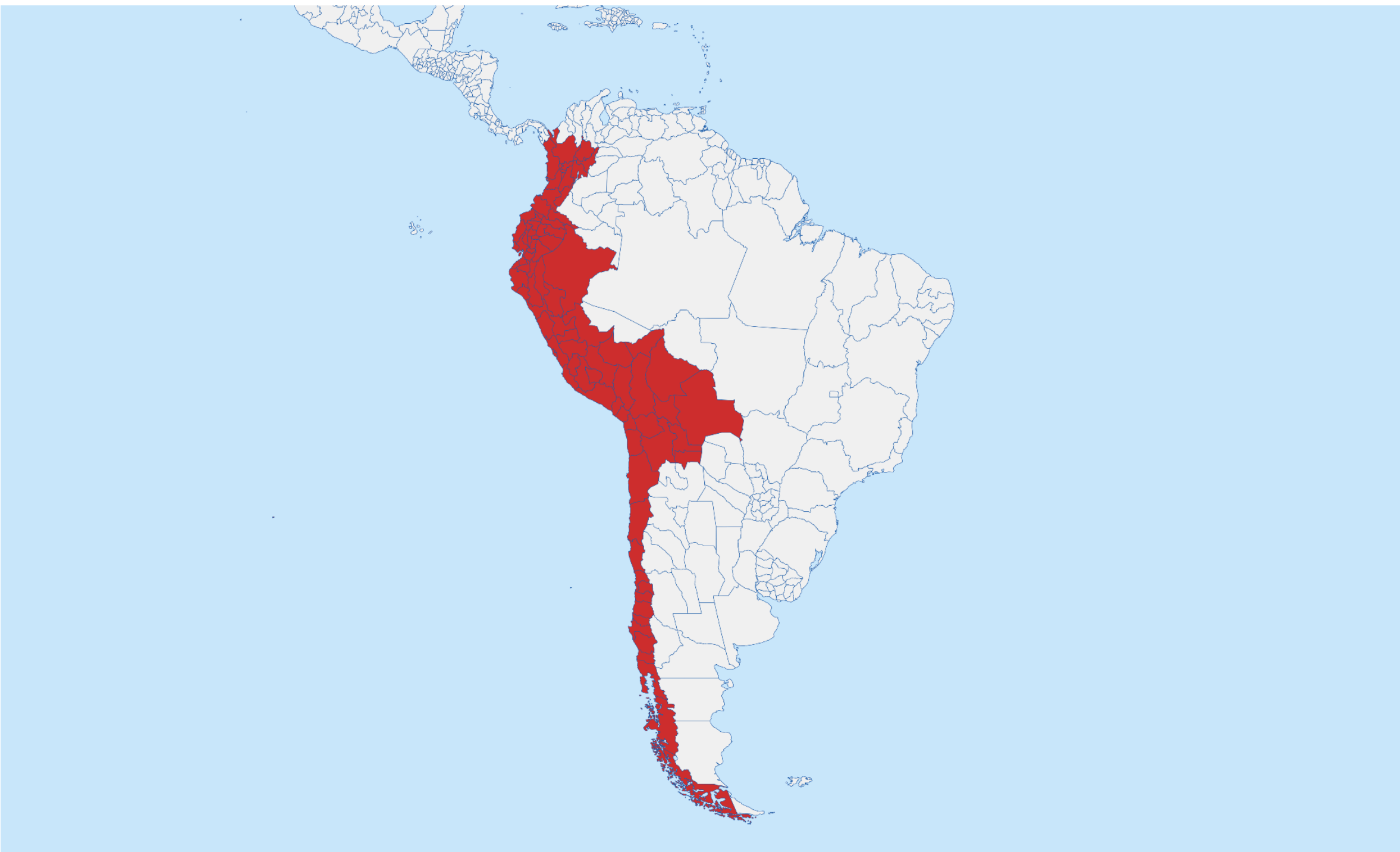

*Figure 13 - Map of Key Areas of Use for Variant 5. Created by the authors.*

## 7. Localization as a Driver of User Experience

The third pillar in our thesis pertains to localization. This analysis has offered an in-depth focus on differences in Spanish throughout Latin America and Spain in order to provide background and reasoning for creating at least 5 separate localized variant models. This leads us to the core question: why implement such changes? The answer is multi-faceted, but it primarily lies in the value of localization as a driver for improved product adoption and engagement. This value is rooted in core marketing practices and ideas, further cemented by the market sizes of Latin America and Spain, not to mention their low-risk market characteristics. Nevertheless, while localization will increase user growth and improve user experience for the 560 million (Wikipedia, 2025) Spanish speakers around the world, there is still another reason why localization matters in this context.

Despite the movement towards globalization within the marketing field, research has found that businesses often find it challenging to fully standardize operations. This is the direct result of a diversity of norms and customs around the world. Differences in consumer needs, conditions of use, purchasing power, culture, and traditions mean that companies must adapt their strategies to suit local markets if they are to achieve relative equal success from nation to nation, and culture to culture. The reality is that strategies are tailored to the singular circumstances of each foreign market (Nguyen, 2016). This approach applies in particular to the present case of Spanish AI model variants. If we view AI models as language-based products then the above data and figures serve to showcase a simple core tenet: that the heterogeneous nature of Latin American countries, and their myriad of differences, from economic characteristics to cultural nuances, makes it impossible for a single standardized Spanish model to suffice for the entire region (Nguyen, 2016).

Perhaps the most effective way to frame this new approach is, rather, to reframe our understanding of the concept of localization itself. It should be seen as an opportunity to foster growth into new, untapped markets, as demonstrated by the studies on the proven decades-long success of Coca-Cola's localization in China (Qing & Zhang, 2023). In the present case, for the approach to Spanish in Latin America, where "globalization means that a product has the 'potential' to be used anywhere; localization is an additional feature added to make it more suitable for use in a 'specific' place" (Qing & Zhang, 2023, p. 408).

# 8. Conclusions and Recommendations

In this paper we have outlined the reasons why localization must be considered a sine qua non component of any growth-oriented AI-mediated product, while also ensuring successful user adoption in the long term. This is true particularly for a language like Spanish, where differences in geographical, contextual, cultural, or linguistic uses and vocabulary can have enormous impacts on a model's output and thus on users' engagement. This is reflected in a number of areas ranging from a model's ability to understand a particular user's intent in context, to its subsequent responses, accuracy, effectiveness, and locale sensitivity. As such, this approach considers a more cost-effective solution to addressing localization, whose primary aim is to avoid the homogenization of linguistic differences. Given the rich depth of variance in Spanish across the Latin America market and Spain, this can, and will, avoid a colonizing approach to sociolinguistic differences across the world. By not implementing a linguistic-based localization approach, we will risk hindering user reach, engagement, and long-term retention, instead of increasing growth in the long term and in a proven low-risk market. Respecting linguistic diversity in Spanish through proper product localization practices will allow hispanophone users' growth across the world to lay down a new blueprint for global marketing in the AI era.

# References

Booth, C. (1993). Miami: the Capital of Latin America. Time Magazine.

Carter, P. M., & Callesano, S. (2018). The social meaning of Spanish in Miami: Dialect perceptions and implications for socioeconomic class, income, and employment. Latin American Studies, 16, 65-90.

Carter, P. M., & Lynch, A. (2015). Multilingual Miami: Current trends in sociolinguistic research. Language and Linguistics Compass, 9(9), 369-385.

Caswell, I. (2022). Google Translate learns 24 new languages. Published internally

Choi, J. K., Elliott, A. R., & Kania, S. (2024). Español práctico: introducción al estudio de la lengua española.

Ethnologue. (2022). Ethnologue: Languages of the World (25th ed.).

Instituto Nacional de Estadística e Informática. (2024). Census Reports. Retrieved from https://www.gob.pe/inei

Lipski, J. M. (2012). Geographical and social varieties of Spanish: An overview. In J. I. Hualde, A. Olarrea, & E. O'Rourke (Eds.), The Handbook of Hispanic Linguistics (1st ed., pp. 1-26). Blackwell Publishing Ltd.

Lipski, J. M. (2018). Dialects of Spanish and Portuguese. In C. Boberg, J. Nerbonne, & D. Watt (Eds.), The Handbook of Dialectology (pp. 498-509). Hoboken, NJ: Wiley.

Merriam-Webster. (2025). Dialect. Retrieved from https://www.merriam-webster.com/dictionary/dialect

Music369.com. (n.d.). Language and vocabulary influences of reggaeton. Retrieved from https://en.music396.com/topic/language-and-vocabulary-influences-of-reggaeton/120182

Nguyen, L. (2016). Standardization versus localization with impacts of cultural patterns on consumption in international marketing. European Journal of Business and Management, 8(35), 140-145.

Olaya Mendoza, K. (2024). Esta es la lengua originaria que más se habla English América Latina: tiene más de 10 millones de hablantes. Noticias de América Latina y el Caribe.

Qing, Z., & Zhang, X. (2023). The degree and generation of localization in marketing: An empirical study based on Coca-Cola's localization strategy. Proceedings of the 2nd International Conference on Business and Policy Studies, 416-417.

Rojas, N. (2022). Reclaiming Indigenous languages, cultures. The Harvard Gazette. Retrieved from https://news.harvard.edu/gazette/story/2022/03/reclaiming-indigenous-languages-cultures/

Tellez, E. S., Moctezuma, D., et al. (2021). Regionalized models for Spanish language variations based on Twitter. Springer Nature.

U.S. Census Bureau. (2023). 2020 Census DHC-A: Hispanic population. Census.gov.

Van Zyl, H., & Meiselman, H. L. (2015). The roles of culture and language in designing emotion lists: Comparing the same language in different English and Spanish speaking countries. Food Quality and Preference, 41, 201-213.

WeChronicle. (n.d.). Exploring the birth, evolution, significance, impact, and influence of reggaeton: Latin urban music and global influence.

Wikipedia. (2024). Demographics of Miami. Retrieved from https://en.wikipedia.org/wiki/Demographics_of_Miami

Wikipedia. (2024). Quechuan languages. Retrieved from https://en.wikipedia.org/wiki/Quechuan_languages

Wikipedia. (2024). Spanish language. Retrieved from https://en.wikipedia.org/wiki/Spanish_language

Wikipedia. (2024). Clitic doubling. Retrieved from https://en.wikipedia.org/wiki/Clitic_doubling

Wikipedia. (2025). List of languages by total number of speakers. Retrieved from https://en.wikipedia.org/wiki/List_of_languages_by_total_number_of_speakers

# Appendix A: Spanish Loanwords from Quechua

Here is a list of some commonly used Spanish loanwords from Quechua. As seen in the table below, words adapted from Quechua are applied in different areas of life, they refer to food, objects, places, people, animals, and are used to express emotions (see Category column). Some words transcend Peru, Ecuador, Colombia, northern Chile and Argentina, which are the regions typically defined as Andean. In addition, these words have different meanings, depending on where they are used.

| Category | Spanish word | Countries | Meaning according to country |
|---|---|---|---|
| Food | *chala* | ARG, BO, CHI, PE, UR | Corn husk |
| Food | *pachamanca* | EC, PE | Traditional dish baked with the aid of hot stones usually under the earth |
| Food | *olluco* | PE | A type of root vegetable |
| Food | *palta* | ARG, CHI, PE, UR | Avocado |
| Food | *papa* | Spanish-speaking countries | Potato |
| | | ARG, MEX, UR | Convenient or easy thing to do |
| | | ARG, UR | Hole, broken |
| | | MEX | Lie |
| | | UR | (colloquially) Beautiful woman |
| Food | *chuño* | ARG, CHI, EC, UR | Potato starch |
| | | BO, PE | Potato starch. / Naturally freeze-dried potato |
| Food | *huacatay* | Spanish-speaking countries | A species of American mint, used as a condiment in some stews |
| Food | *choclo* | ARG, BO, CHI, CO, EC, PAR, PE | Corn |
| Expression | *arrarray* | EC | Used for the expression of heat. It is used to express the sensation of heat or burning |
| Expression | *achachay* | EC, COL | Used to express the feeling of cold |
| | | PE | Used to express fear |
| Expression | *atatay* | BO | Used to denote pain |
| | | EC | Used to express the feeling of disgust |
| Expression | *ayayay* | PE | Used to express various feelings, especially those of grief and pain |
| Expression | *amarcar* | EC | Take in your arms or sponsor a child |
| Expression | *yapa* | ARG, CHI, EC, PAR, PE, UR | Addition, especially that which is given as a tip or gift. / Gratuitously |
| | | BO | Gratuitously |
| Object | *cancha* | Spanish-speaking countries | Space for the practice of certain sports or shows (sports center, court) / Skill acquired through experience |
| | | ARG, CHI, COL, EC, El Salv, MEX, PAN, PAR, PE, Dom. Rep. | Land, space, premises or flat and unburdened site |
| | | ARG, CHI, EC, PAR, PE, Dom. Rep. | A spacious corral or fence to deposit certain objects. Wooden court. |
| | | Cuba, PAR | Hippodrome |
| | | EC | A place where the course of a river is wider and clearer |
| | | COL, PAR | The amount charged by the owner of a gambling house |
| | | ARG, BOL, EC, El Salv., MEX, PAN, PAR, UR, VE | (colloquially) Used to ask others to make way |
| Object | *chapa* | EC | Name for door lock (universal) / Police officer |
| | | ARG | License plate |
| | | NIC, PAN | Denture |
| | | PE | Nickname/ Bottle cap/Door lock |
| Object | *pirca* | ARG, CHI, EC, PE | Dry-stone wall |
| Object | *chullo* | PE | Hat with earmuffs to protect against the cold |
| Animal | *llama* | Spanish-speaking countries | Llama |
| Animal | *puma* | Spanish-speaking countries | Puma |
| Culture | *Pachamama* | PE | Pachamama (goddess of Earth and fertility) |
| Place | Machu Picchu | Spanish-speaking countries | Machu Picchu (Inca citadel) |
| Place | *pampa* | Spanish-speaking countries | Extensive grassy plain |
| Place | *puna* | Spanish-speaking countries | High treeless plateau in the Andes |
| Food/Object | *chaucha* | ARG | Green bean / Pod / A small amount of money |
| | | PAR | Green bean |
| | | UR | Green bean / Pod / (colloquially) Annoying or boring, because of its poor quality |
| | | BO | Small silver or nickel coin. / A small amount of money |
| | | CHI | Small silver or nickel coin. / Early or small potato that is left for seed. / A small amount of money |
| People | *ñaño, ña* | BO, EC | Close friend / Brother or sister |
| | | PE | Close friend |
| | | COL | Spoiled |
| | | PAN | Spoiled / Homosexual |
| | | ARG | Brother or sister |
| | | PE | Child |
| People / Object | *guagua* | ARG, BO, CHI, COL, EC, PE, CU | Kid / Sweet bread in the shape of a child |
| | | Cuba | bus |

# Appendix B: Indigenous Languages in Latin America

"The 10 Latin America countries with the highest number of indigenous languages"

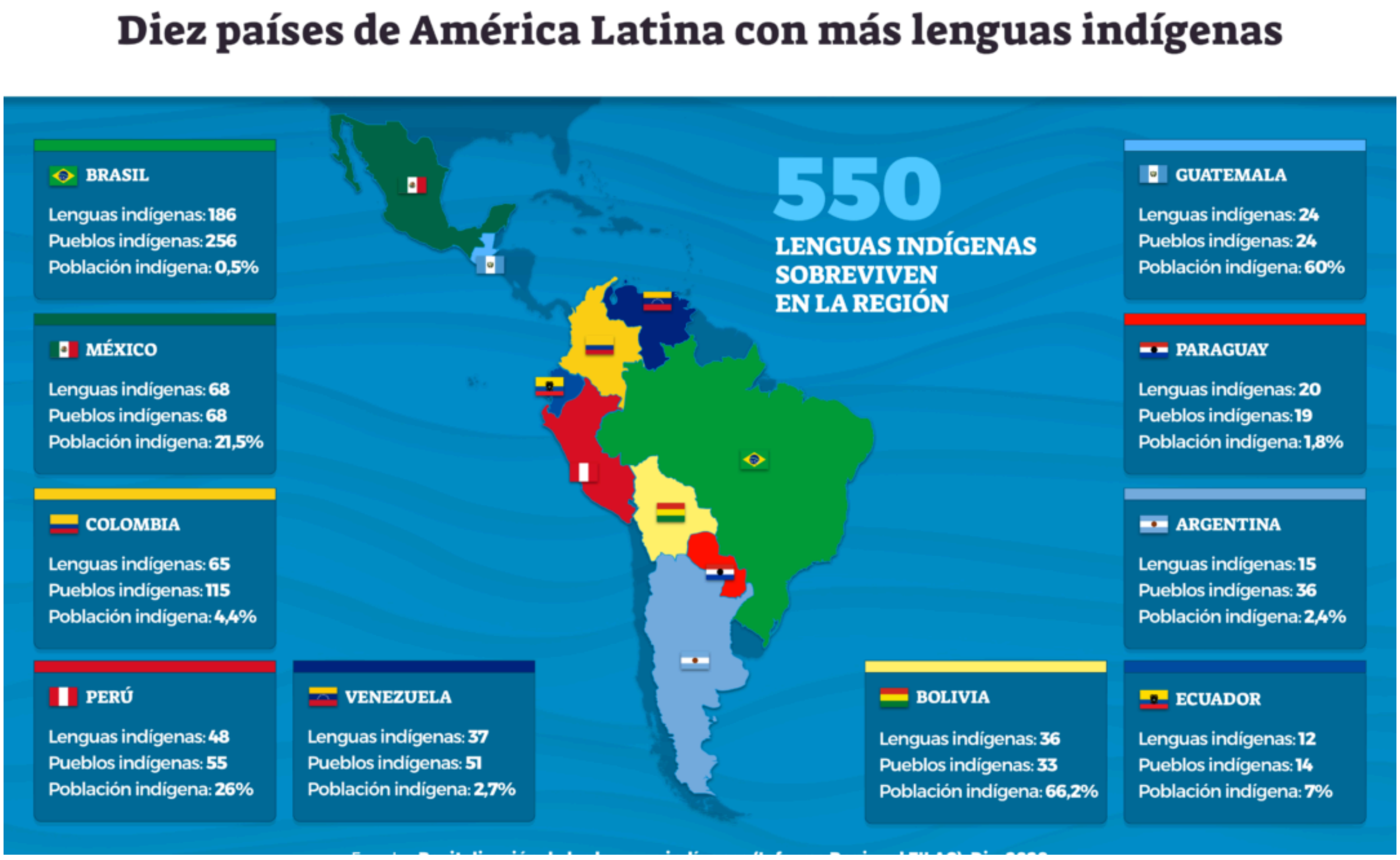

This figure shows 550 indigenous languages that are active today in the Latin American region. For each country the information includes: Number of indigenous languages, number of recognized indigenous peoples, and what percentage of the total population of the country is indigenous. For example, in the box in the upper-left corner, which corresponds to Brazil, there are 186 indigenous languages, 256 indigenous peoples, and that represents 0.5% of the total population. For each country in the table the same information is provided in that order.

Source: Multiple Eds. (2022). "*Patrimonio": Un tercio de las lenguas indígenas de América Latina y el Caribe están en peligro de desaparecer*." (One third of indigenous languages in Latin America and the Caribbean are in danger of disappearing). Retrieved from: https://uta.pressbooks.pub/espanolpractico/chapter/capitulo-6-espanol-en-contacto-con-otras-lenguas/